\documentclass{article}

\usepackage[final]{new_in_ML}
\usepackage{graphicx}

\usepackage[utf8]{inputenc} 
\usepackage[T1]{fontenc}    
\usepackage{hyperref}       
\usepackage{url}            
\usepackage{booktabs}       
\usepackage{amsfonts}       
\usepackage{nicefrac}       
\usepackage{microtype}      
\usepackage{xcolor}         

\title{CLAMP: A Contrastive Language And Molecule Pre-training 
Network}

\author{Neel Redkar $^{1}$ \\
$^1$University of California Los Angeles\\
\texttt{\{Email:neel.redkar@gmail.com\}}}

\begin{document}

\maketitle

\begin{abstract}
This paper highlights a shift in how to approach material generation. Instead of material-to-material, we propose a language-to-material generation architecture that utilizes millions of untapped data points. Using a web scraper to collect crystal text pairs from open-source research papers, a contrastive model can be trained using a convolutional graph neural network encoder and a language encoder. This would allow unsupervised zero-shot classification which can be trained by taking advantage of linguistic structure. Without any specific training data, an ~82\% accuracy was achieved and ~75\% accuracy for photocatalyst prediction with an extremely small dataset. This novel network could ideally be cross-applied to any reaction that can be described via text, opening completely new methods to think about 3D chemical framework generation. In the full experiment diffusion models would likely be incorporated to fully exploit the latent space.
\end{abstract}

\section{Introduction}
\subsection{Metal-Organic Frameworks (MOFs)}
Metal-Organic Frameworks are organo-metals joined by organic ligands that can can have an assortment of properties. The simple 3-dimensional structure leads to them being candidates for a litany of processes—artificial photosynthesis, carbon capture, water treatment, efficient electrolysis, and more. Their extremely high surface area (porosity) also make them promising choices for solid adsorbants of CO\textsubscript{2}. Artificial photosynthesis \& CO\textsubscript{2} adsorbance is particularity interesting due to the energy implications, but data is a limiting factor.

\subsection{Machine Learning Architectures}
The MOF space is diverse with many possible properties, which makes exploration key to finding specific crystals with desired attributes. Current methods are highly reliant on experimentation where they need to guess and test possible MOFs. This leads to staying close to the known, as well as often missing radically novel MOFs that could function better. Machine learning is a solution which is efficient at taking large feature spaces and searching for maxima. This has been a common method for finding novel MOFs for different processes \cite{chong_applications_2020}. 

Common methods utilized are Monte Carlo trees and generative adversarial neural networks, both of which use large amounts of data—10K+ \cite{chong_applications_2020} \cite{zhang_machine_2021}. These are examples of high throughput screening—common with MOFs. This approach uses large amounts of data, which makes it impossible because experimental data is costly and in low amounts.

Monte Carlo trees (MCT) are usually promising in such tasks, but the MOF linearization to fit a tree loses essential data that can hurt end products \cite{zhang_machine_2021}. Spatial properties are not accounted for in both MCT \& adversarial methods which lead to losses in accuracy.

The largest flaw though is that either the dataset is large and the model overfits the domain to only make small changes, or more commonly the dataset is small and the crystals generated are not plausible. A middle ground between creativity and accuracy hasn't yet been discovered. If solved, we could see large improvements in MOF quality due to rapid \& diverse experimentation.

\section{Proof of Concept Experiments}
\subsection{Data Gathering}
In the initial phase of the experiment, web scrapers were employed to extract key descriptors about each metal-organic framework (MOF). To further enhance the extracted information, a tree recursive summarization model basd on the Pegasus Summarization Large Language Model was required to condense the abstracts into concise descriptions.

In addition to extracting text data, CIFs (crystallography information files) were downloaded from the Crystallography Open Database and examined for atomic composition and structural integrity. This dataset ranges is around 222k crystal descriptor pairs, which is extremely large compared to the general 500 pairs used for MOF regression models. Key to unsupervised learning, this enables a litany of unsupervised models to be used to extract more accurate \& deal with the larger noise of language.

This is a fundamentally different dataset, with text to crystal pairs, compared to general crystal experimental values—opening the route to natural language techniques with crystals. To our knowledge this is the first dataset of this kind, and the dataset generation scripts can be scaled up even more with funding \& license contributions.

\begin{figure*}
    \centering
    \begin{minipage}{0.45\textwidth}
        \centering
        \label{fig:clamp}
        \includegraphics[width=1.2\textwidth]{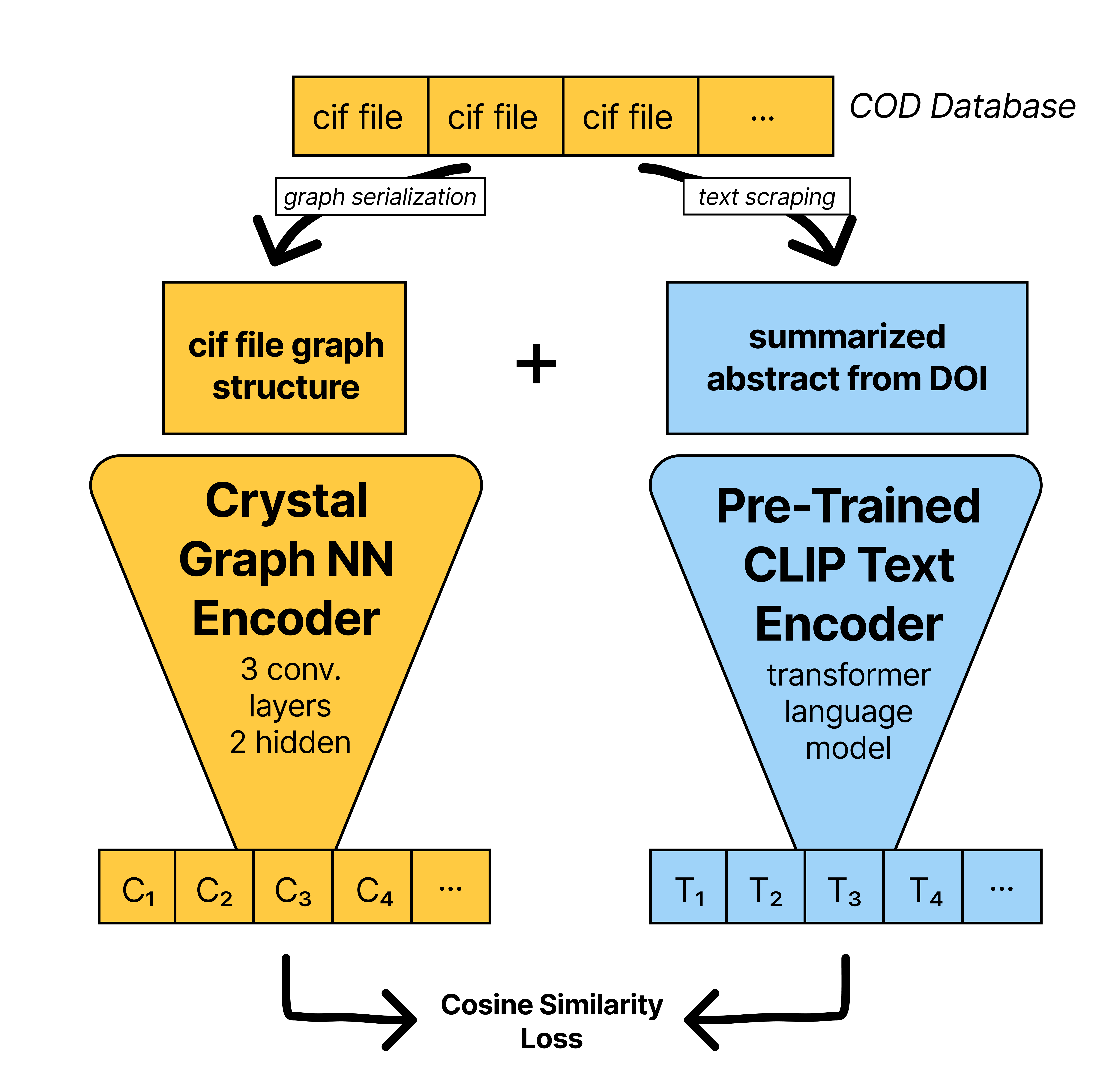} 
        \caption{CLAMP model architecture diagram. Left: CGCNN, Right: CLIP Text Encoder}
    \end{minipage}\hfill
    \begin{minipage}{0.45\textwidth}
        \centering
        \label{fig:zero_shot}
        \includegraphics[width=1.2\textwidth]{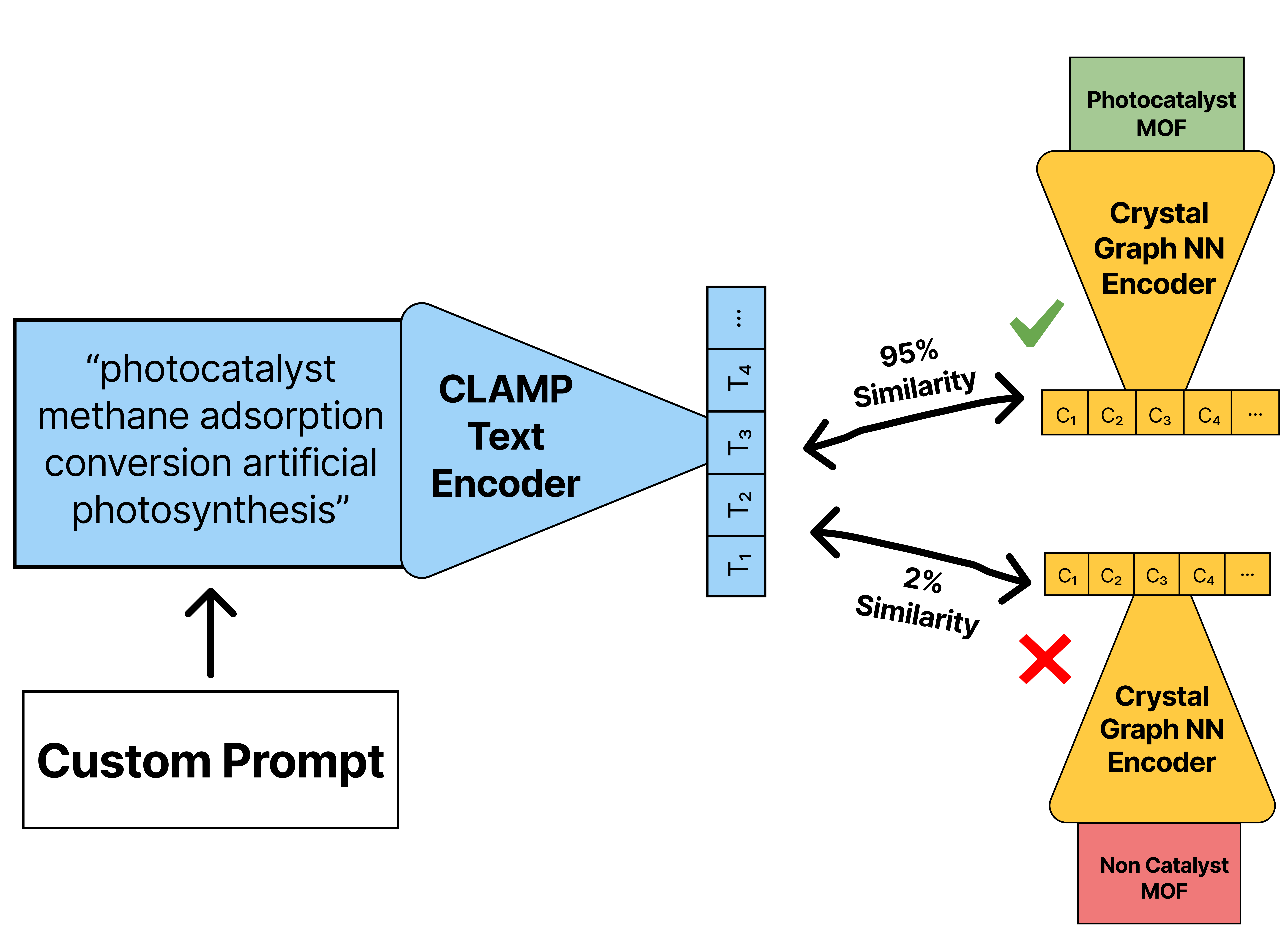} 
        \caption{Zero shot classification process, which can be utilized with a custom prompt. Future work includes integration with prior diffusion models \cite{wu2022diffusionbased}. }
    \end{minipage}
\end{figure*}
\subsection{Contrastive Language Molecule Architecture}
A contrastive network was built on a subset of the data, 72k out of 222k because of computational constraints, to compare the crystal-text pairs. The language model utilized transfer learning from the CLIP ViT text transformer \cite{radford_learning_2021}. A SOTA prior understanding of grammar would benefit the model rather than random weights. By using CLIP priors, there is a possibility of adding new encoders that are built on top of CLIP with high compatability—crystal to image. The weights for the model were not frozen during training time to speed up convergence with lower compute.

For the crystal encoder, a Crystal Graph Convolutional Neural Network (CGCNN) was utilized \cite{xie_crystal_2018} to model spatial properties of crystals. Data efficiency wise, CGCNN's are able to achieve high accuracies with ~500 data points. By expanding the amount of data by multiple magnitudes, the larger task of embedding was predicted to work well. By utilizing a convolutional graph network there is also greater communication between nodes, allowing simulation of spatial reactions between different atoms.

Cosine similarity was utilized as the inter-batch loss function to maximize similarity of pairings, while maintaining unique embeddings. This was mainly used due to its success with the pretrained CLIP objective but also to prefer small divergences over large \cite{radford_learning_2021}.

\subsection{Zero Shot Classification}
For zero shot classification embeddings were compared to given text embeddings, and the similarity was calculated. Then classification labels were selected by the embedding vectors with the greatest similarity between areas. This has two uses 1) Give quantifiable data on accuracy of the embedding model 2) Allow novel reward functions for RL generation for qualitative data, something impossible before.

Around 14k crystals were used as validation data to the 72k crystal pairs used due to computational constraints. On the general testing dataset a 82\% accuracy was achieved in matching text to crystal pairs in classification. This was extremely successful in a general sense which proves that the model has generalized to many different tasks. Through review, 12 specific photocatalysts were selected used in zero shot classification with a 75\% accuracy was achieved. This shows the model was able to comprehend natural language and utilize it in context specific domains.

\section{Conclusion \& Diffusion Future Work}
In experimental design the model architechture seems to work as a proof of concept. With more testing this novel graph contrastive network could be expanded to the full 222k datapoints collected—possibly resulting in highly improved accuracies. Key contributions include 1) A first of its kind large scale dataset (222k points) of crystal text pairs, allowing for natural language integration into material generation 2) A contrastive model (trained on 72k out of the 222k due to computational constraints) to create an embedding space similar to CLIP for other models to build off of. Both are extremely important to possibly highlight a new paradigm of language to material generation—key for low data tasks which are extremely common. Dataset \& source code are open source at \href{https://github.com/neelr/clamp}{https://github.com/neelr/clamp}.

Objectives can be utilized with molecular diffusion from prior information presented at NeurIPS 2022 \cite{wu2022diffusionbased}. The current model uses noise to generate diffusion vectors for the crystal, but CLAMP embeddings could be utilized to guide it towards a text objective. This would help in bridging the natural language barrier in material generation from small changes to large scale generation—similar to what DALLE3 did for generation.

This paper provides a glimpse into a possible world where artificial photosynthesis, carbon capture, water treatment, and so much more are at the tips of our fingers. MOFs could be generated to capture carbon dioxide, qualitative data like bandgaps could be quantified, and avenues could open for fully zero shot material generation. 
\bibliographystyle{plain}
\bibliography{main}
\end{document}